\documentclass[a4,a4wide]{article}
\usepackage{aaai}
\usepackage{times}
\usepackage{helvet}
\usepackage{courier}
\usepackage{times}
\usepackage{latexsym}
\usepackage{amssymb}
\usepackage{xspace}
\usepackage{listings}
\usepackage{url}
\usepackage{graphicx}
\usepackage{epsfig}
\usepackage{amsmath}
\usepackage{times}
\usepackage[ruled,linesnumbered,vlined]{algorithm2e}
\usepackage{subfigure}

 \usepackage{times}
 \usepackage{epic}
 \usepackage{eepic}
 \usepackage{epsfig}
 \usepackage{url}
 \usepackage{xspace}
 \usepackage{latexsym}
 \usepackage{alltt}
 \usepackage{listings}
 \usepackage{mdwlist}

\newif\ifreport

\newcommand{\nop}[1]{}

\newcommand{\inco}{\ensuremath{\mathit{Incoherent}}\xspace}

\newcommand{\set}[1]{\{ #1 \}}

\newcommand{\derives}{\mbox{\,\texttt{:\hspace{-0.15em}-}}\,\xspace}

\newcommand{\R}{\ensuremath{r}}

\newcommand{\HR}{\ensuremath{H(\R)}}
\newcommand{\BR}{\ensuremath{B(\R)}}
\newcommand{\BpR}{\ensuremath{B^+(\R)}}

\def\naf{\ensuremath{\raise.17ex\hbox{\ensuremath{\scriptstyle\mathtt{\sim}}}}\xspace}

\newcommand{\dlv}{{\sc DLV}\xspace}

\newcommand{\posbody}[1]{\ensuremath{B^+(#1)}}
\newcommand{\negbody}[1]{\ensuremath{B^-(#1)}}

\newenvironment{simpleprogram}[1][]
   {\vspace{-0.5ex}\begin{itemize}\item[]
      \tt
      \begin{tabbing}
      \code{#1}\ \= \kill
   }
   {\end{tabbing}\end{itemize}\vspace{-3ex}}

\newenvironment{simplealignedprogramstub}[1][]
   {\vspace{-0ex}%\begin{itemize}\item[]
      \begin{tabbing}
      #1\kill
   }
   {\end{tabbing}
\vspace{-6ex}}

\newenvironment{sublabeledprogram}[1][]
   {\begin{array}{ll}\setlength{\arraycolsep}{0pt}}
   {\end{array}}

\newcommand{\code}[1]{\ensuremath{#1}}

\newcommand{\U}{{\cal A}}

\renewcommand{\P}{\mathcal{P}}

\newcommand{\wasptwo}{WASP 2.0\xspace}
\newcommand{\wasp}{WASP 1.0\xspace}
\newcommand{\clasp}{clasp\xspace}

\newcommand{\cmodels}{Cmodels3\xspace}
\newcommand{\minisat}{MiniSAT\xspace}

\begin{document}
\nocopyright
\title{Preliminary Report on \wasptwo
\thanks{
% by project PIA KnowRex POR FESR 2007- 2013 BURC n. 49 s.s. n. 1 16/12/2010,
%by MIUR project FRAME PON01\_02477/4, and
This research has been partly supported by the European Commission, European Social Fund of Regione Calabria,
the Regione Calabria under project PIA KnowRex POR FESR 2007- 2013 BURC n. 49 s.s. n. 1 16/12/2010, and 
the Italian Ministry of University and Research under PON project ``Ba2Know S.I.-LAB'' n. PON03PE\_0001.
}
}
\author{Mario Alviano, Carmine Dodaro and Francesco Ricca\\
Department of Mathematics and Computer Science, University of Calabria, Italy\\
\url{{alviano,dodaro,ricca}@mat.unical.it}\\
}
\maketitle
\begin{abstract}
\begin{quote}
Answer Set Programming (ASP) is a declarative programming paradigm.
% which
%has been proposed in the area of non-monotonic reasoning and logic programming.
The intrinsic complexity of the evaluation of ASP programs makes the development 
of more effective and faster systems a challenging research topic.
%
%The traditional evaluation flow of an ASP program requires to combine an instantiator and a solver.
%The instantiator produces a propositional program equivalent to the input one, 
%which is evaluated by a solver implementing a backtracking search algorithm.
%
This paper reports on the recent improvements of the ASP solver WASP.
WASP is undergoing a refactoring process which will end up in the release 
of a new and more performant version of the software.
In particular the paper focus on the improvements to the core evaluation algorithms working on normal programs.
A preliminary experiment on benchmarks from the 3rd ASP competition belonging to the NP class is reported. 
The previous version of WASP was often not competitive with alternative solutions on this class.
The new version of WASP shows a substantial increase in performance.
\end{quote}
\end{abstract}

\section{Introduction}\label{sec:introduction}

Answer Set Programming (ASP)~\cite{gelf-lifs-91} is a declarative programming paradigm which
has been proposed in the area of non-monotonic reasoning and logic programming.
The idea of ASP is to represent a given computational
problem by a logic program whose answer sets correspond to solutions,
and then use a solver to find them.

%The ASP language considered here allows disjunction in rule heads and
%nonmonotonic negation in rule bodies. These features make ASP very expressive; 
%all problems in the second level of the polynomial hierarchy are indeed
%expressible in ASP \cite{eite-etal-97f}. 
%Therefore, ASP is strictly more expressive than SAT (unless $P=NP$).
Despite the intrinsic complexity of the evaluation of ASP, 
after twenty years of research many efficient ASP systems have been developed.
(e.g.~\cite{alvi-etal-2011-dl2,gebs-etal-2007-ijcai,lier-mara-2004-lpnmr}).
The availability of robust implementations made ASP a powerful tool for developing advanced applications
in the areas of Artificial Intelligence, Information Integration, and Knowledge Management.
%for example, ASP has been used in industrial applications \cite{grass-etal-2009-apps}, and for team-building \cite{ricca-etal-2012-tplp},
%semantic-based information extraction \cite{manna-etal-2012-tldks}, and e-tourism \cite{ricca-etal-2010-fi}.
These applications of ASP have confirmed the viability of the use of ASP.
Nonetheless, the interest in developing more effective and faster systems is still a crucial and challenging 
research topic, as witnessed by the results of the ASP Competition series (see e.g.~\cite{DBLP:journals/tplp/CalimeriIR14}).%

This paper reports on the recent improvements of the ASP solver for propositional programs WASP~\cite{DBLP:conf/lpnmr/AlvianoDFLR13}.
The new version of WASP is inspired by several techniques that were originally introduced for SAT solving, 
like the Davis-Putnam-Logemann-Loveland (DPLL) backtracking search algorithm~\cite{davi-etal-62}, 
{\em clause learning}~\cite{zhan-etal-2001}, {\em backjumping}~\cite{gasc-79}, 
{\em restarts}~\cite{gome-etal-98}, and {\em conflict-driven heuristics}~\cite{mosk-etal-2001}.
The mentioned SAT-solving methods have been adapted and combined with state-of-the-art pruning techniques 
adopted by modern native ASP solvers~\cite{alvi-etal-2011-dl2,gebs-etal-2007-ijcai}. 
In particular, the role of Boolean Constraint Propagation in SAT-solvers is taken by a procedure 
combining the {\em unit propagation} inference rule with inference techniques based on ASP
program properties. In particular, support inferences are implemented via Clark's completion, and 
the implementation of the well-founded operator is based on source pointers~\cite{simo-etal-2002}.

In the following, we overview the techniques implemented by the 2.0 version of WASP,
focusing on the improvements to the core evaluation algorithms working on normal programs.
Then we compare the new implementation with the previous one.

We also report on a preliminary experiment in which we compare the old and new versions of WASP
with the latest version of clasp, which is the solver that won the 3rd and 4th edition of the ASP competition.
Benchmarks were taken from the 3rd ASP competition and belong to the NP class, 
i.e., the class of problems where the previous version of WASP was often not competitive with alternative solutions.
The result show that \wasptwo is substantially faster than \wasp and is often competitive with clasp.

%%%%%%%%%%%%%%%%%%%%%%%%%%%%%%%%%%%%%%%%%%%%%%%%%%%%%%%%%%%%%%%%%%%%%%%%%%%
\section{ASP Language}\label{sec:preliminaries}

Let $\U$ be a countable set of propositional atoms.
A {\em literal} is either an atom (a positive literal), or an atom preceded by the {\em
  negation as failure} symbol $\naf$ (a negative literal).
The complement of a literal $\ell$ is denoted $\overline{\ell}$, i.e., 
$\overline{a} = \naf a$ and $\overline{\naf a} = a$ for an atom $a$.
This notation extends to sets of literals, i.e.,
$\overline{L} := \{\overline{\ell} \mid \ell \in L\}$ for a set of literals $L$.

A \emph{program} is a finite set of rules of the following form:
\begin{equation}\label{eq:rule}
    a_0 \derives a_1, \ldots, a_m, \naf a_{m+1}, \ldots, \naf a_n  
\end{equation}
where $n\geq m\geq 0$ and each $a_i$ ($i = 0, \ldots, n$) is an atom.
The atom $a_0$ is called head,
and the conjunction 
$a_1, \ldots, a_m, \naf a_{m+1}, \ldots, \naf a_n$ 
is referred to as body.
Rule $r$ is said to be
regular if $\HR \neq \bot$,
where $\bot$ is a fixed atom in $\U$,
and a constraint otherwise.
For a rule $r$ of the form (\ref{eq:rule}), 
the following notation is also used:
$\HR$ denotes the head atom $a_0$;
$\BR$ denotes the set $\{a_1, \ldots, a_m, \naf a_{m+1}, \ldots, \naf a_n\}$ of body literals;
$\posbody{\R}$ and $\negbody{\R}$ denote
the set of atoms appearing in positive and negative body literals, respectively;
$C(r) := H(r) \cup \overline{B(r)}$ is the clause representation of $r$.

An \emph{interpretation} $I$ is a set of literals, i.e., $I \subseteq \U \cup \overline{\U}$.
Intuitively, literals in $I$ are true, literals whose complements are in $I$ are false, and all other literals are undefined.
$I$ is total if there are no undefined literals,
and $I$ is inconsistent if $\bot \in I$ or there is $a \in \U$ such that $\{a, \naf a\} \subseteq I$.
An interpretation $I$ satisfies a rule $r$ if $C(r) \cap I \neq \emptyset$,
while $I$ violates $r$ if $C(r) \subseteq \overline{I}$. 
A \emph{model} of a program $\P$ is a consistent, total interpretation satisfying all rules of $\P$. 
The semantics of a program $\P$ is given by the set of its \emph{answer sets}
(or stable models) \cite{gelf-lifs-91},
where an interpretation $I$ is an answer set for $\P$ if 
$I$ is a subset-minimal model of the reduct $\P^I$
obtained by deleting from $\P$ each rule $\R$ such that $\negbody{\R} \cap I \neq \emptyset$,
and then by removing all the negative literals from the remaining rules.

\section{Answer Set Computation in \wasptwo}\label{sec:modelgenerator}
In this section we review the algorithms implemented in \wasptwo.
The presentation is properly simplified to focus on the main principles.

\subsection{Completion and Program Simplification}
The first step of the evaluation in \wasptwo is a program transformation step.
The input program first undergoes a Clark's completion transformation step, and 
then is simplified applying techniques in the style of satelite \cite{DBLP:conf/sat/EenB05}.
%
%Given an atom $a \in \U$, let $Heads(a)$ be the set of rules where $a$ is the head atom, i.e., $Heads(a) := \{r \mid r \in \P, H(r) = a\}$.
Given a rule $r \in \P$, let $aux_r$ denote a fresh atom, i.e., an atom not appearing elsewhere.
The completion of $\P$, denoted $Comp(\P)$, consists of the following clauses:
\begin{itemize}
\item $\{\naf a, aux_{r_1}, \ldots, aux_{r_n}\}$ for each atom $a$ occurring in $\P$, where $r_1,\ldots,r_n$ are the rules of $\P$ whose head is $a$;
\item $\{H(r), \naf aux_r\}$ and $\{aux_r\} \cup \overline{B(r)}$ for each rule $r \in \P$;
\item $\{\naf aux_r, \ell$\} for each $r \in \P$ and $\ell \in B(r)$.
\end{itemize}
After the computation of Clark's completion, \emph{simplification} techniques are applied \cite{DBLP:conf/sat/EenB05}.
These consist of polynomial algorithms for strengthening and for removing redundant clauses, and also include atoms elimination by means of clause rewriting.

\subsection{Main Algorithm}
An answer set of a given propositional program $Comp(\P)$ is computed in \wasptwo by
using Algorithm~\ref{alg:mg}, which is similar to the DPLL procedure in SAT solvers.
Initially, interpretation $I$ is set to $\{\naf \bot\}$.
Function Propagate (line~\ref{ln:alg:propagate}) extends $I$ with those literals that can be deterministically inferred.
This function returns false if an inconsistency (or conflict) is detected, true otherwise. 
%(A more detailed description of this function is reported in section Propagation.)
When no inconsistency is detected, interpretation $I$ is returned if total (lines~2--3).
Otherwise, an undefined literal, say $\ell$, is chosen according to some heuristic criterion (line~5).
Then computation then proceeds with a recursive call 
to ComputeAnswerSet on $I \cup \{\ell\}$ (line~\ref{ln:alg:recursive}).
In case the recursive call returns an answer set, the computation ends returning it (lines~\ref{ln:alg:isAS}--\ref{ln:alg:returnAS}).
Otherwise, the algorithm unrolls choices until consistency of $I$ is restored (backjumping; lines~\ref{ln:alg:isInco}--\ref{ln:alg:returnInco}),
and the computation resumes by propagating the consequences of the clause
learned by the conflict analysis.  
Conflicts detected during propagation are analyzed by procedure {\em AnalyzeConflictAndLearnClauses} (line~\ref{ln:alg:analyze}).

The main algorithm is usually complemented with some heuristic techniques
that control the number of learned clauses (which may be exponential in number),
and possibly restart the computation to explore different branches of the search tree.
Moreover, a crucial role is played by the heuristic criteria used for selecting branching literals.
\wasptwo adopts the same branching and deletion heuristics of the SAT solver \minisat~\cite{een-etal-2003}.
The restart policy is based on the sequence of thresholds introduced in \cite{luby-etal-93}.

Propagation and clause learning are described in more detail in the following.

\begin{algorithm}[t]
 \SetKwInOut{Input}{Input}
 \SetKwInOut{Output}{Output}
 \Input{An interpretation $I$ for a program $Comp(\P)$}
 \Output{An answer set for $Comp(\P)$ or \inco}
\Begin{
		\While{\emph{Propagate($I$)}\label{ln:alg:propagate}}
		{
			\If{$I$ is total\label{ln:alg:total}}
			{
				\Return{$I$};\\
			}
			$\ell$ := ChooseUndefinedLiteral();\label{ln:alg:choose}\\
			$I'$ := ComputeAnswerSet($I \cup \{\ell\}$)\;\label{ln:alg:recursive}
			\If{$I' \neq \inco$\label{ln:alg:isAS}}
			{
			    \Return{$I'$}\;\label{ln:alg:returnAS}
			}
	   		 \If{there are violated (learned) clauses\label{ln:alg:isInco}}
		    {
		        \Return{\inco}\;\label{ln:alg:returnInco}
		    }
		}
		
		AnalyzeConflictAndLearnClauses($I$);\label{ln:alg:analyze}\\
		\Return{\inco}\;
}
\caption{Compute Answer Set}\label{alg:mg}
\end{algorithm}

\begin{function}[t]
    \While{\emph{UnitPropagation($I$)}}
    {
        \If{\emph{\textbf{not} WellFoundedPropagation}($I$)}
        {
	        \Return{true};\\
        }
    }
    \Return{false}\;
    \caption{Propagate($I$)}\label{fn:propagate}
\end{function}

\paragraph{Propagation.}\label{sec:propagation}
\wasptwo implements two deterministic inference rules for pruning the search
space during answer set computation. These propagation rules are named
\emph{unit} and \emph{well-founded}. %, and are described in the following.
Unit propagation is applied first (line~1 of function Propagate).
It returns false if an inconsistency arises.
Otherwise, well-founded propagation is applied (line~2).
Function WellFoundedPropagation may learn an implicit clause in $P$, in which case true is returned and unit propagation is applied on the new clause.
When no new clause can be learned by WellFoundedPropagation, function Propagate returns true to report that no inconsistency has been detected.
More in details, unit propagation is as in SAT solvers:
An undefined literal $\ell$ is inferred by unit propagation if there is a rule $r$
that can be satisfied only by $\ell$, i.e., $r$ is such that $\ell \in C(r)$ and $C(r) \setminus \{\ell\} \subseteq \overline{I}$.
Concerning well-founded propagation, we must first introduce the notion of unfounded set.
A set $X$ of atoms is \emph{unfounded} if 
for each rule $r$ such that $H(r) \cap X \neq \emptyset$, at least one of the following conditions is satisfied:
(i) $\overline{B(r)} \cap I \neq \emptyset$; (ii) $\BpR \cap X \neq \emptyset$; (iii) $I \cap \HR \setminus X  \neq \emptyset$.
Intuitively, atoms in $X$ can have support only by themselves.
When an unfounded set $X$ is found, function WellFoundedPropagation learns a clause forcing falsity of an atom in $X$.
Clauses for other atoms in $X$ will be learned on subsequent calls to the function, unless an inconsistency arises during unit propagation.
In case of inconsistencies, indeed, the unfounded set $X$ is recomputed.

\paragraph{Conflict Analysis and Learning.}\label{sec:learning}
Clause learning acquires information from conflicts 
in order to avoid exploring the same search branch several times. 
%
%\paragraph{Learning from Propagation.} 
%In this case, 
\wasptwo adopts a learning schema based on the concept of the first Unique Implication Point (UIP) \cite{mosk-etal-2001}, which is computed by analyzing the so-called implication graph.
Roughly, the implication graph contains a node for each literal in $I$, and arcs from $\ell_i$ to $\ell_0$ ($i = 1,\ldots,n$; $n \geq 1$) if literal $\ell_0$ is inferred by unit propagation on clause $\{\ell_0,\ldots,\ell_n\}$.
Each literal $\ell \in I$ is associated with a {\em decision level}, 
corresponding to the depth nesting level of the recursive call to ComputeAnswerSet on which $\ell$ is added to $I$.
A node $n$ in the implication graph is a UIP for a decision level $d$ if all paths 
from the choice of level $d$ to the conflict literals pass through $n$. 
The first UIP is the UIP for the decision level of the conflict that is closest to the conflict.
The learning schema is as follows:
Let $u$ be the first UIP. Let $L$ be the set of literals different form $u$ 
occurring in a path from $u$ to the conflict literals. 
The learned clause comprises $u$ and each literal $\ell$ such that 
the decision level of $\ell$ is lower than the one of $u$ and 
there is an arc $(\overline{\ell}, \ell')$ in the implication graph for some $\ell'\in L $.

\subsection{Comparing \wasp and \wasptwo}
In this section we compare \wasptwo to \wasp. 
First of all we observe that \wasp does not implement any program transformation phase, 
whereas \wasptwo applies both Clark's completion and program simplification in the style of~\cite{DBLP:conf/sat/EenB05}.
The addition of this preprocessing step brings advantages in both terms of simplifying the implementation of
the propagation procedure and in terms performance.
The Clark's completion introduces a number of clauses that represent support propagation, 
which is implemented natively in \wasp instead. The subsequent program simplification step 
optimizes the program by eliminating redundant atoms (also introduced by the completion) 
and shrinking definitions. This results in a program that is usually easier to evaluate.
Concerning the well-founded operator both \wasptwo and \wasp compute 
unfounded sets according to the \emph{source pointers}~\cite{simo-etal-2002} technique.
\wasp, which implements a native inference rule, immediately infers unfounded atoms as false, 
and updates a special implementation of the implication graph. In contrast, \wasptwo 
learns a clause representing the inference (also called loop formula) and propagates
it with unit propagation. This choice combined with Clark's completion allows to 
simplify conflict analysis, learning and backjumping. Indeed, \wasp implements specialized 
variants of these procedures that require the usage of complex data structures that are difficult to optimize.
Since in \wasptwo literals are always inferred by the UnitPropagation procedure, 
we could adopt an implementation of these strategies optimized as in modern SAT solvers.
Finally both \wasptwo and \wasp implement conflict-driven branching heuristics.
\wasptwo uses a branching heuristic inspired to the one of \minisat, 
while \wasp uses an extension of the BerkMin~\cite{gold-novi-02} heuristics extended by adding 
a look-ahead technique and an additional ASP-specific criterion.

%%%%%%%%%%%%%%%%%%%%%%%%%%%%%%%%%%%%%%%%%%%%%%%
\section{Experiment}\label{sec:experiments}

\begin{table*}
	\caption{Average running time and number of solved instances} %\vspace{0.1cm}
	\label{tab:competition}
	\centering
	\begin{tabular}{lrr|rrr|rrr|rrr}
	\multicolumn{3}{c}{} 		 			&\multicolumn{3}{|c}{\bf{\clasp}} 			&\multicolumn{3}{|c}{\bf{\wasp}} 				&\multicolumn{3}{|c}{\bf{\wasptwo}} \\
	{\bf Problem}					& \bf{\#\ } & \bf{\#$_{all}$\ } 		& \bf{sol.} 	& \multicolumn{1}{c}{\bf{t}} 			&\multicolumn{1}{c|}{\bf{t$_{all}$}} 	&\bf{sol.} 		& \multicolumn{1}{c}{\bf{t}}			& \multicolumn{1}{c|}{\bf{t$_{all}$}} 	&\bf{sol.} 		& \multicolumn{1}{c}{\bf{t}}	 		& \multicolumn{1}{c}{\bf{t$_{all}$}} \\
\cline{1-12}
DisjunctiveScheduling 			& 10 	& 5	& 5 		& 16.8 	& 16.8 	& 5 		& 29.0 	& 29.0 	& 5 		& 188.4 	& 188.4		\\
GraphColouring 				& 10 	& 3	& 4 		& 88.0 	& 20.6 	& 3 		& 50.5 	& 50.5 	& 3 		& 3.3 	& 3.3		\\
HanoiTower 					& 10 	& 2	& 7 		& 126.0 	& 49.8 	& 2 		& 214.0 	& 214.0 	& 7 		& 52.5 	& 18.3		\\
KnightTour 					& 10 	& 6	& 10 	& 14.3 	& 0.3 	& 6 		& 93.5 	& 93.5 	& 10 	& 16.0 	& 0.6		\\
Labyrinth 						& 10 	& 8	& 9 		& 74.4 	& 74.7 	& 8		& 118.7 	& 118.7 	& 10 	& 85.8 	& 84.7		\\
MazeGeneration 				& 10 	& 10	& 10 	& 0.3 	& 0.3 	& 10 	& 19.9 	& 19.9 	& 10 	& 2.7 	& 2.7		\\
MultiContextSystemQuerying 		& 10 	& 10	& 10 	& 5.1 	& 5.1 	& 10 	& 122.4 	& 122.4 	& 10 	& 9.4 	& 9.4		\\
Numberlink 					& 10 	& 6	& 8 		& 21.1 	& 0.6 	& 6 		& 24.3 	& 24.3 	& 7 		& 8.7 	& 5.5		\\
PackingProblem 				& 10 	& 0	& 0 		& - 		& - 		& 0 		& - 		& - 		& 0 		& - 		& -			\\
SokobanDecision 				& 10		& 5	& 10 	& 101.5 	& 2.8	& 5 		& 212.8 	& 212.8 	& 7 		& 97.8 	& 14.4		\\
Solitaire 						& 10 	& 2	& 2 		& 124.9 	& 124.9 	& 3 		& 183.1 	& 198.0 	& 4 		& 8.7 	& 6.0		\\
WeightAssignmentTree 			& 10 	& 1	& 5 		& 119.2 	& 22.4 	& 1 		& 297.3 	& 297.3 	& 3 		& 282.3 	& 97.9		\\
\cline{1-12}
{\bf Total}					& {\bf 120} & {\bf 58} &	{\bf 80}	& {\bf 62.9}	& {\bf 20.5}	& {\bf 59}		& {\bf 124.1}	& {\bf 95.6}	& {\bf 76}		& {\bf 68.7}	& {\bf 34.6}		\\
    \end{tabular}
\end{table*}

In this section we report the results of an experiment assessing the performance of \wasptwo.
In particular, we compare \wasptwo with \wasp and \clasp. All the solvers used gringo 3.0.5 \cite{DBLP:conf/lpnmr/GebserKKS11} as grounder.
\clasp and \wasp has been executed with the same heuristic setting used in \cite{DBLP:conf/lpnmr/AlvianoDFLR13}. Concerning \clasp we used the version 3.0.1.
The experiment was run on a Mac Pro equipped with two 3 GHz Intel Xeon X5365 (quad core) processors, 
with 4 MB of L2 cache and 16 GB of RAM, running Debian Linux 7.3 (kernel ver. 3.2.0-4-amd64). 
Binaries were generated with the GNU C++ compiler 4.7.3-4 shipped by Debian.
Time limit was set to 600 seconds. Performance was measured using the tools pyrunlim and pyrunner ({\em https://github.com/alviano/python}).

Tested instances are among those in the System Track of the 3rd ASP Competition~\cite{DBLP:journals/tplp/CalimeriIR14}, in particular all instances in the NP category.
This category includes planning domains, temporal and 
spatial scheduling problems, combinatorial puzzles, graph problems, and
a number of real-world domains in which ASP has been applied.
(See \cite{DBLP:journals/tplp/CalimeriIR14} for an exhaustive description of the benchmarks.)

Table~\ref{tab:competition} summarizes the number of solved instances and the average running times in seconds for each
solver.
In particular, the first two columns report the total number of instances (\#) and the number of instances that are solved by all solvers (\#$_{all}$), respectively; the remaining columns report the number of solved instances within the time-out (sol.), and the running times averaged both over solved instances ($t$) and over instances solved by all variants ($t_{all}$). 

We observe that \wasptwo outperforms \wasp.
In fact, \wasptwo solved 17 instances more than \wasp, and also the improvement on the average execution time is sensible, with a percentage gain of around 64\% on instances solved by all systems.
On the other hand, \clasp is faster than \wasptwo, with a percentage gain of around 41  \% on the same instances.
Moreover, \clasp solved 4 instances more than \wasptwo.

Analyzing the results in more detail, there are some specific benchmarks where \wasptwo and \clasp exhibit significantly performances.
Two of these problems are SokobanDecision and WeightAssignmentTree, where \clasp solved 3 and 2 instances more than \wasptwo, respectively, while \wasptwo solved 2 instances more than \clasp in Solitaire.
We also note that the performance of WASP deteriored in DisjunctiveScheduling.
This is due to the initial steps of the computation, and in particular to the simplification procedure, which in this case removes 80\% of clauses and 99\% of atoms.
However, there are cases in which simplifications play a crucial role to improve performance of the answer set search procedure.
For example, in HanoiTower, where \wasptwo performs better than other systems, more than half of the variables are removed in a few seconds.

%%%%%%%%%%%%%%%%%%%%%%%%%%%%%%%%%%%%%%%%%%%%%%%%%%%%%%%%
\section{Related Work}\label{sec:related}
\wasp is inspired by several techniques used in SAT solving that were first introduced for Constraint Satisfaction and QBF solving.

Some of these techniques were already adapted in non-disjunctive ASP solvers like Smodels$_{cc}$~\cite{ward-schl-2004-lpnmr},
\clasp~\cite{gebs-etal-2007-ijcai}, Smodels~\cite{simo-etal-2002}, \cmodels~\cite{lier-mara-2004-lpnmr},
and \dlv~\cite{ricc-etal-2006-aicom}.
More in detail, \wasptwo differs from \cmodels~\cite{lier-mara-2004-lpnmr} that are based 
on a rewriting into a propositional formula and an external SAT solver.
\wasptwo differs from \dlv~\cite{alvi-etal-2011-dl2} and the Smodels variants,
which features a native implementation of all inference rules.
Our new solver is more similar to \clasp, but there are differences concerning the restart policy, constraint deletion and branching heuristics.
\wasptwo adopts as default a policy based on the sequence of thresholds introduced in \cite{luby-etal-93},
whereas \clasp employs by default a different policy based on geometric series.
Concerning deletion of learned constraints, \wasptwo adopts a criterion inspired by \minisat, while \clasp implements a technique introduced in Glucose \cite{DBLP:conf/ijcai/AudemardS09}.
%Nonetheless, the program can grow in \clasp up to three times the size of the original input, while \wasp does not limit the growth of the program.
Moreover, \clasp adopts a branching heuristic based on BerkMin~\cite{gold-novi-02}
with a variant of the MOMS criterion which estimates the effect of the candidate literals in short clauses.

%%%%%%%%%%%%%%%%%%%%%%%%%%%%%%%%%%%%%%%%%%%%%%%%%%%%%%%%%
\section{Conclusion}\label{sec:conclusion}
In this paper we reported on the recent improvement of the ASP solver \wasp.
We described the main improvements on the evaluation procedure
focusing on the improvements to the core evaluation algorithms working on normal programs.
The new solver was compared with both its predecessor and the latest version of clasp
on on benchmarks belonging to the NP class, where \wasp was not competitive.
The result is very encouraging, since \wasptwo improves substantially w.r.t. \wasp and is often competitive with clasp.

Future work concerns the reengineering of disjunctive rules, aggregates, and weak constraints,
as well as the introduction of a native implementation of choice rules. 

\bibliographystyle{aaai}

\end{document}